\documentclass[final,12pt,twocolumn]{elsarticle}
\usepackage[margin=1.5cm,includefoot]{geometry}
\usepackage[table,xcdraw]{xcolor}
\usepackage[refcheck=false,todonotes=false]{autopaper}
\usepackage{array}
\newcolumntype{P}[1]{>{\centering\arraybackslash}p{#1}}

\usepackage{listings}

\definecolor{codegreen}{rgb}{0,0.6,0}
\definecolor{codegray}{rgb}{0.5,0.5,0.5}
\definecolor{codepurple}{rgb}{0.58,0,0.82}
\definecolor{backcolour}{rgb}{0.95,0.95,0.92}

\lstdefinestyle{mystyle}{
    backgroundcolor=\color{backcolour},   
    commentstyle=\color{codegreen},
    keywordstyle=\color{magenta},
    numberstyle=\tiny\color{codegray},
    stringstyle=\color{codepurple},
    basicstyle=\ttfamily\scriptsize,
    breakatwhitespace=false,         
    breaklines=true,                 
    captionpos=b,                    
    keepspaces=true,                                  
    showspaces=false,                
    showstringspaces=false,
    showtabs=false,                  
    tabsize=2,
    floatplacement=C,
    framexleftmargin=2pt,
    framexrightmargin=2pt,
    framextopmargin=1pt,
    framexbottommargin=1pt,
    frame=tb, framerule=0pt,
}

\usepackage{makecell}  % Include this in the preamble for splitting text into multiple lines
\usepackage{adjustbox} % Include this in the preamble for angled headers
\usepackage{tabularx} % Include this in the preamble
\usepackage{pifont}   % Include this in the preamble for check marks
\newcommand{\cmark}{\ding{51}} % Check mark
\usepackage{CJKutf8}
\usepackage[utf8]{inputenc}

\usepackage{booktabs}

\begin{document}

%to deal with white space better
\sloppy

% title, authors, abstract, keywords
\begin{frontmatter}

\title{Honegumi: An Interface for Accelerating the Adoption of Bayesian Optimization in the Experimental Sciences}
\author[utah,acceleration]{Sterling G.~Baird}\ead{sterling.baird@utoronto.ca}
\author[utah]{Andrew R. Falkowski}\ead{andrew.falkowski@utah.edu}
\author[utah,acceleration]{Taylor D.~Sparks}\ead{sparks@eng.utah.edu}

\address[utah]{Department of Materials Science and Engineering, University of Utah, Salt Lake City, UT 84108, USA}
\address[acceleration]{Acceleration Consortium, University of Toronto.\ 80 St George St, Toronto, ON M5S 3H6}

\date{August 2024}

\begin{abstract}
    \noindent Bayesian optimization (BO) has emerged as a powerful tool for guiding experimental design and decision-making in various scientific fields, including materials science, chemistry, and biology. However, despite its growing popularity, the complexity of existing BO libraries and the steep learning curve associated with them can deter researchers who are not well-versed in machine learning or programming. To address this barrier, we introduce Honegumi (\begin{CJK}{UTF8}{min}骨組み\end{CJK}), a user-friendly, interactive tool designed to simplify the process of creating advanced Bayesian optimization scripts. Honegumi offers a dynamic selection grid that allows users to configure key parameters of their optimization tasks, generating ready-to-use, unit-tested Python scripts tailored to their specific needs. Accompanying the interface is a comprehensive suite of tutorials that provide both conceptual and practical guidance, bridging the gap between theoretical understanding and practical implementation. Built on top of the Ax platform, Honegumi leverages the power of existing state-of-the-art libraries while restructuring the user experience to make advanced BO techniques more accessible to experimental researchers. By lowering the barrier to entry and providing educational resources, Honegumi aims to accelerate the adoption of advanced Bayesian optimization methods across various domains.
\end{abstract}


\end{frontmatter}

\section{Introduction}

Bayesian optimization (BO) has become a critical tool in guiding experimental and computational research in both lab and production settings. In contrast to traditional design of experiments approaches, the core BO loop sees a probabilistic model iteratively fit to observed data such that it models and eventually predicts the optimal region of a design space. The physical sciences, especially materials, chemistry, and biology have benefited greatly with numerous published case studies using the technique \cite{strieth-kalthoff_delocalized_2024, shields2021bayesian, agarwal2021discovery, chen2020bayesian, hickman2022bayesian}. %Need to add more famous case studies here.
Growing interest in self-driving laboratories has emphasized the need for automated experimental design policies to govern decision making. BO is and will continue to play a large role in this space \cite{arroyave2022perspective, tom2024self}. % Self-driving Labs for Chemistry and Materials Science (chemical reviews)
The core concepts and extensions of BO have been extensively reviewed within the literature, which we refer the unfamiliar reader to in the interest of brevity \cite{frazier_tutorial_2018, garnett_bayesian_2023, shahriari_taking_2016}.  

% How to make this less antagonistic towards Ax and Botorch?
Despite growing interest, there remains a barrier to entry in many BO libraries that dissuade researchers who aren't well versed in machine learning. Understanding the concepts behind BO does not immediately prepare a researcher to interact with many of the existing BO libraries such as Ax and Botorch \cite{balandat_botorch_2020}, which often employ library specific conventions and code structures in addition to making assumptions under the hood that aren't immediately clear to new users. Accessing more advanced features in these libraries typically requires extensive knowledge of the library codebase, which can limit researchers wanting to take advantage of new developments in the field of BO. Furthermore, these libraries typically aren't oriented towards experimental research, resulting in limited crossover between tutorial content and experimentalist needs. Building more targeted libraries is one solution, but runs the risk of minimal maintenance in the absence of larger organization support and may not include newer state of the art optimization tools.

Here we present Honegumi (\begin{CJK}{UTF8}{min}骨組み\end{CJK}), pronounced "ho-nay-goo-mee," a tool for interactively creating minimal working examples for advanced Bayesian optimization topics. Honegumi translates to "skeletal framework," which the developed tool echoes in its templating-oriented structure. Taking the form of a selection grid, the Honegumi interface dynamically constructs unit-tested python scripts based on user selections that can then be modified to meet user needs. Accompanying this selection tool is a suite of conceptual and coding tutorials that inform practitioners on the practical aspects of BO and guide them in making these modifications.

\section{Honegumi Overview}

\begin{figure*}
    \centering
    \includegraphics[width=0.9\textwidth]{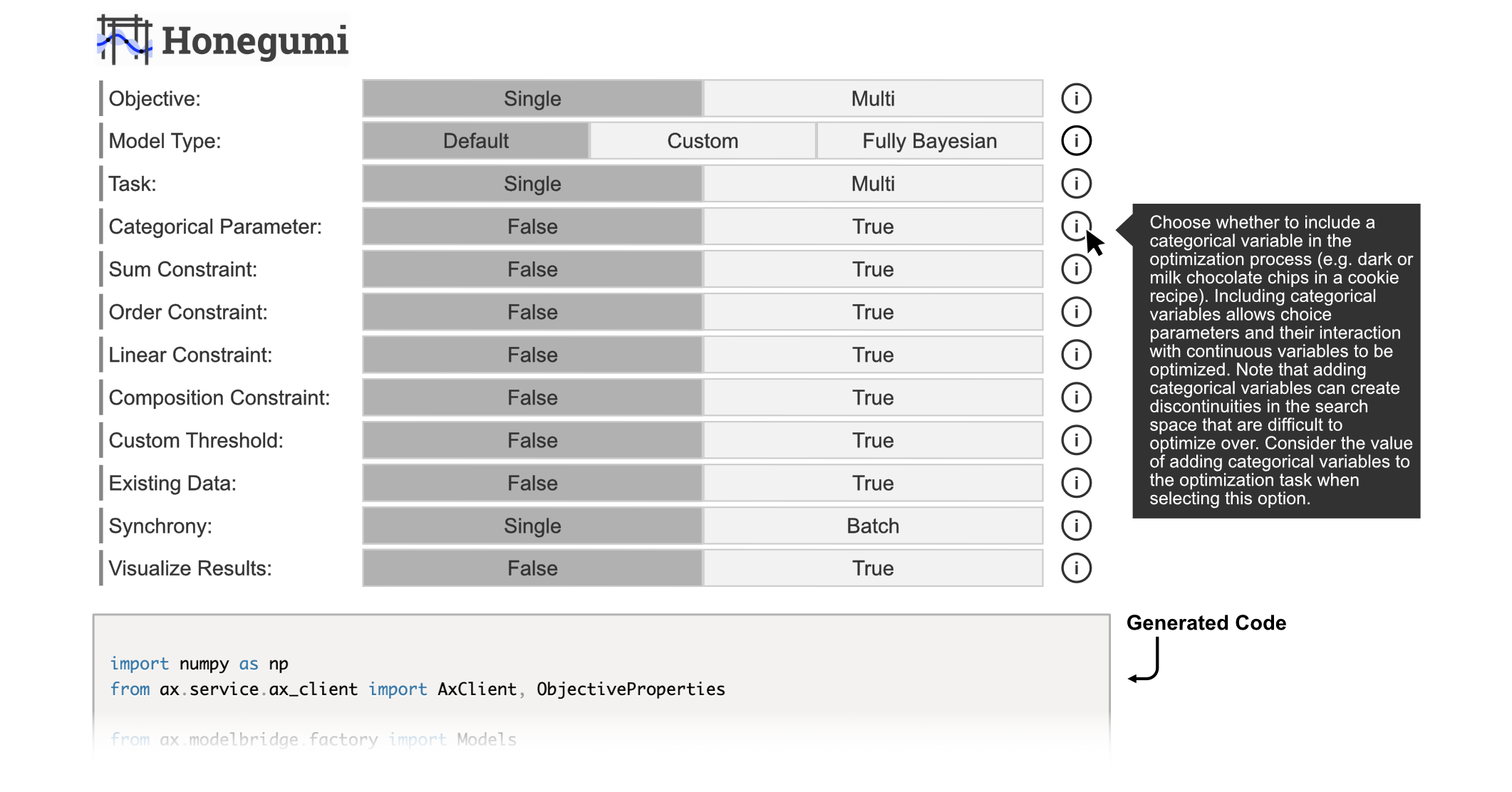}
    \caption{Default selection interface available on the Honegumi web page with highlighted tooltip information.}
    \label{fig:selection_interface}
\end{figure*}

The Honegumi interface takes the form of a selection grid that allows users to configure optimization scripts through a series of toggleable options. Each row defines a distinct parameter or setting; for example, whether to construct a "Single" or "Multi" objective optimization script. The order of options reflects a balance between the order in which one proceeds in defining an optimization task, similarity in function, and the location within the constructed script. Tooltips provide additional context for each row and guide users in making appropriate selections for their application. An image of the interface is shown in \cref{fig:selection_interface}.

% ARF: Ax doesn't have a publication associated with it.

% ARF: Not sure if this is the right word here. Trying to emphasize that Ax tutorials all concern simple benchmarking problems with arbitrary constraints applied to them. Practical applications of BO are typically quite different and a little more messy.

Honegumi is built on top of the Adaptive Experimentation Platform (Ax Platform) developed by Meta. Ax provides a high-level tool set for performing BO campaigns and pulls from the state-of-the-art optimization and Gaussian process libraries BoTorch \cite{balandat_botorch_2020} and GPytorch \cite{gardner_gpytorch_2018}. A consequence of building on top of many libraries is that interacting with the codebase can be challenging. While a plethora of well written tutorials are present on the Ax website, these are typically oriented towards benchmark tasks and frequently span several operational modalities of the Ax framework (Loop, Service, and Developer), which can hinder researchers who may need to pull from several tutorials to find something that meets their specific needs. The Honegumi interface, on the other hand, simplifies and restructures how users access the powerful tools and state-of-the-art models built into Ax by providing a consistent framework that helps users learn the library's core features and extend them to their application needs. All scripts conform to the Ax Service API, which strikes a balance between simplicity and flexibility and is the recommended means of engagement with the Ax Platform. 

A Python script is dynamically constructed based on the selected options below the selection interface. This script functions as a minimum working BO example that can be copied and run in a code editor. Constructed tests are unit-tested to ensure functionality across the 4096 unique combinations of settings. This makes it easy for researchers to engage with the library's features without worrying about low-level implementation bugs. The standardized structure of the outputs provides a stable foundation of best practices for learning and expanding on simple BO scripts.

\section{An Educational Curriculum}

To guide users in modifying the scripts generated by Honegumi, a suite of conceptual and coding tutorials are included on the Honegumi website. The conceptual tutorials seek to emphasize important and often nuanced BO concepts and guide users in designing a script for their particular problem and avoiding certain pitfalls. The coding tutorials present a materials or chemistry problem and walk users through the process of translating a problem statement into proper Honegumi selections and code modifications. These are explained in more detail in the sections below.

\subsection{Concept Tutorials}

\begin{figure*}[h!]
    \centering
    \includegraphics[width=\textwidth]{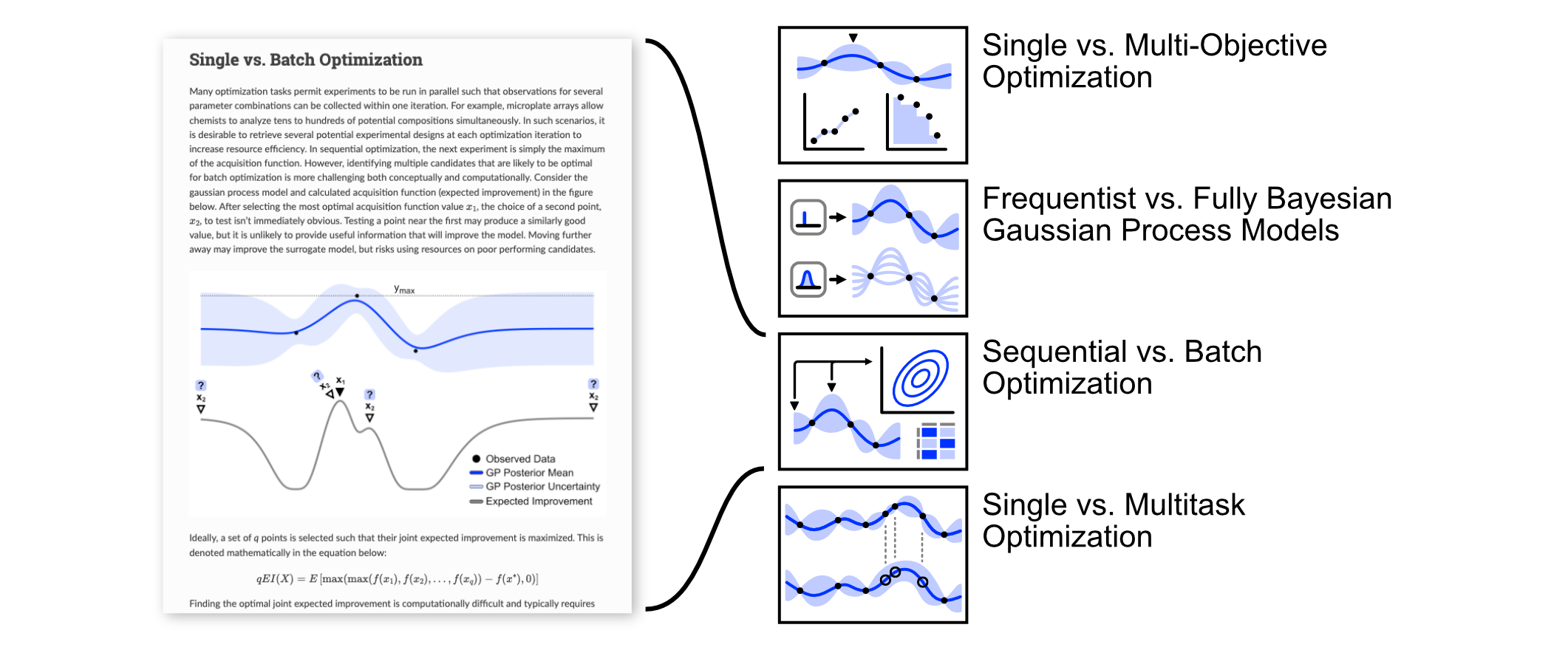}
    \caption{List of concept tutorials and an example of the formatting.}
    \label{fig:concept_tutorials}
\end{figure*}

There is a risk in using tools with which one is unfamiliar, and BO can go astray if applied inappropriately. To further guide researchers, we provide several conceptual tutorials that expand on several topics that often cause confusion or are misapplied. The conceptual tutorials do not seek to be rigorous, but rather to provide practical understandings of nuanced topics through the use of written explanation, visuals, and mathematical equations where appropriate. The aim is to inform researchers such that they can make selections with greater confidence and understand the shortcomings of different optimization approaches. At the time of publication, there are four concept tutorials available on the Honegumi webpage covering single versus multi-objective optimization, batch optimization, standard versus fully Bayesian Gaussian process models, and the advantages and disadvantages of multitask BO. These are visualized along with an example of a tutorial in \cref{fig:concept_tutorials}. 

At the end of each concept tutorial is a link to a coding tutorial that shows how this concept can be applied using Honegumi, giving users the opportunity to see it in action within Ax. Additionally, a list of relevant references, publications, videos, and blogs that provide deeper insights on the concept are provided for those seeking deeper understanding. Honegumi does not seek to be a replacement for a proper education in BO, but rather to bridge the knowledge gap and make it easier for researchers to expand their knowledge of BO and engage with deeper literature.

\subsection{Coding Tutorials}

A variety of coding tutorials are provided on the Honegumi website that translate common experimental challenges into functional optimization scripts. All coding tutorials follow a common structure that proceeds from an experimental problem statement inspired by a real world application, the selections in Honegumi one might make given this problem statement, modification of the Honegumi constructed framework, to visualizations and analysis of results. While taking the guise of real experimental scenarios, the objective functions provided in each coding tutorial are synthetic and are intended to be functional examples rather than challenges. Each coding tutorial ends with a list of possible next steps and ways of further expanding the script. This inclusion is pedagogical and seeks to expand user's grasp of the tool and knowledge of what is possible with the tool set. The list of coding tutorials along with the concept(s) they explore is presented in \cref{tab:coding-tutorials}.

\begin{table*}[!ht]
\renewcommand{\arraystretch}{0.15}
\centering
\caption{List of coding tutorials and explored optimization concepts. "Constr." refers to constraint. "obj." refers to objective. "thresh." refers to threshold.}
\label{tab:coding-tutorials}
% \begin{tabularx}{\textwidth}{@{}P{3.155in}*{12}{p{0.405cm}}@{}}
\begin{tabularx}{\textwidth}{@{}P{2.55in}*{12}{p{0.405cm}}@{}}
\textbf{Tutorial Name} &
  \rotatebox{55}{\textbf{Multi-Obj.}} &
  \rotatebox{55}{\textbf{Custom Model}} &
  \rotatebox{55}{\textbf{Multi-Task}} &
  \rotatebox{55}{\textbf{Categorical Vars.}} &
  \rotatebox{55}{\textbf{Sum Constraints}} &
  \rotatebox{55}{\textbf{Order Constraint}} &
  \rotatebox{55}{\textbf{Linear Constraint}} &
  \rotatebox{55}{\textbf{Composition}} &
  \rotatebox{55}{\textbf{Custom Thresh.}} &
  \rotatebox{55}{\textbf{Historical Data}} &
  \rotatebox{55}{\textbf{Batching}} &
  \rotatebox{55}{\textbf{Visualization}} \\ \midrule
\end{tabularx}
\begin{tabularx}{\textwidth}{@{}l*{12}{|P{0.39cm}}|@{}}
\makecell[l]{Optimizing 3D Printed Material\\[-2pt] Strength Under Constraints}  &   &   &   &  \cellcolor{gray!15}\cmark &  &   & \cellcolor{gray!15}\cmark &   &   &   &   & \cellcolor{gray!15}\cmark \\ \\
\makecell[l]{Optimizing a Polymer Compound \\[-2pt] for Strength and Density}    & \cellcolor{gray!15}\cmark &   &   &   &  &   &   & \cellcolor{gray!15}\cmark & \cellcolor{gray!15}\cmark & \cellcolor{gray!15}\cmark &   & \cellcolor{gray!15}\cmark \\ \\
\makecell[l]{Optimizing Anti-Corrosion Coatings\\[-2pt] with Noisy Measurements} &   & \cellcolor{gray!15}\cmark &   &   &  & \cellcolor{gray!15}\cmark &   & \cellcolor{gray!15}\cmark &   &   & \cellcolor{gray!15}\cmark & \cellcolor{gray!15}\cmark \\ \\
\makecell[l]{Optimizing MAX Phases\\[-2pt] with Featurization}                   &   & \cellcolor{gray!15}\cmark &   &   &  &   &   &   &   & \cellcolor{gray!15}\cmark &   & \cellcolor{gray!15}\cmark \\ \\
\makecell[l]{Multi-Task Optimization Across\\[-2pt] Ceramic Binder Systems}      &   & \cellcolor{gray!15}\cmark & \cellcolor{gray!15}\cmark &   &  &   &   &   &   &   &   & \cellcolor{gray!15}\cmark \\ \\
\makecell[l]{Benchmarking Across Several \\[-2pt] Acquisition Functions}         & \cellcolor{gray!15}\cmark &   &   &   &  &   &   &   &   &   &   &  \\
\bottomrule
\end{tabularx}
\end{table*}

\renewcommand{\arraystretch}{1.0}

\section{Under the Hood}

\begin{figure}[!b]
    \centering
    \includegraphics[width=\linewidth]{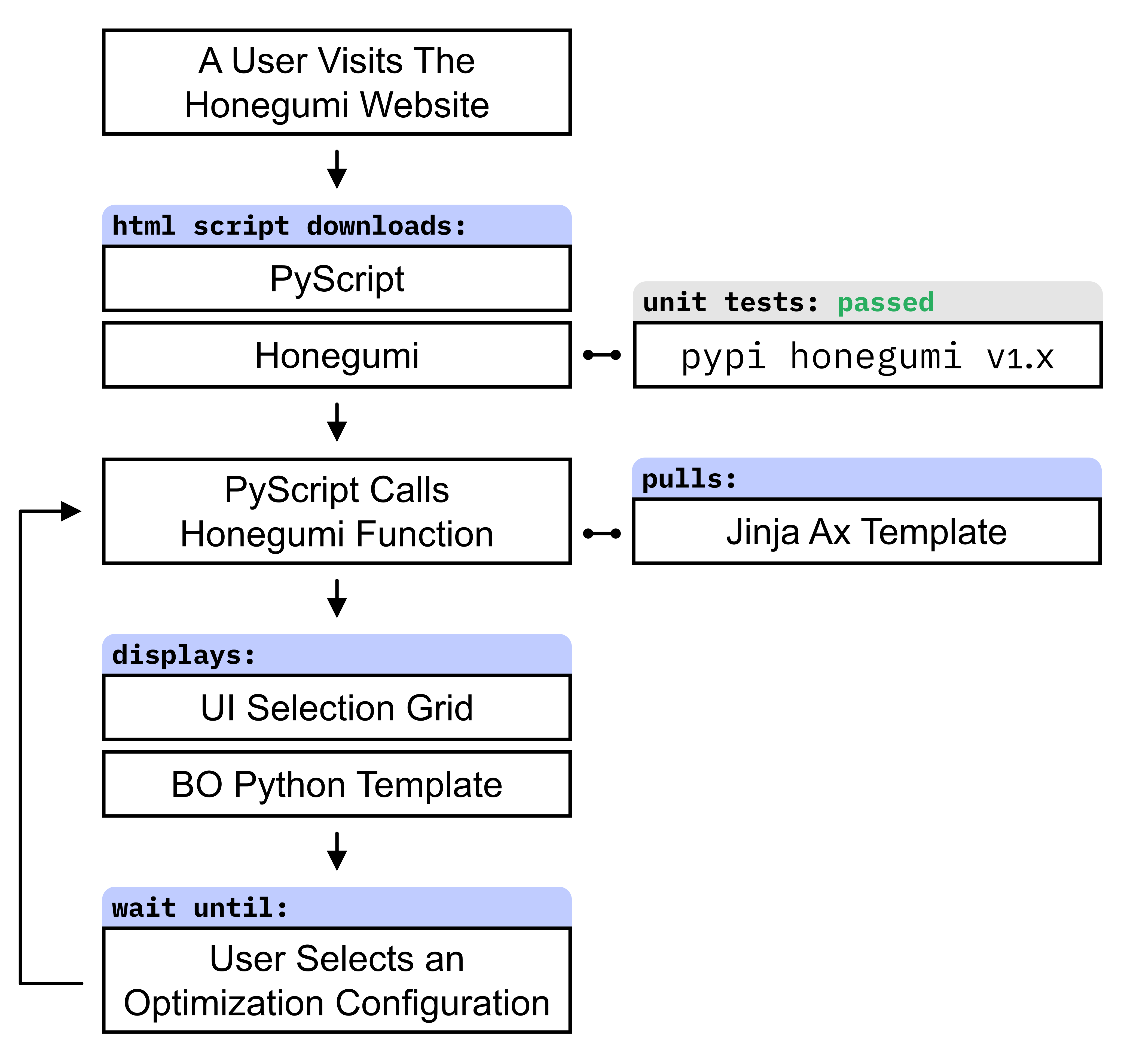}
    \caption{Flowchart detailing the honegumi workflow.}
    \label{fig:flowchart}
\end{figure}

Behind the scenes, a suite of connected tools facilitates the assembly of coding templates, testing of relevant combinatorial selection options, and presentation of the interface to the user. There are two key tools that enable this are \href{https://jinja.palletsprojects.com/}{Jinja} and \href{https://pyscript.net/}{PyScript}, described below. Jinja is a Python-friendly templating engine that encodes the logic of the selection interface. PyScript is a package enabling Python in the browser, which allows the Honegumi package to be run on a ReadtheDocs instance. The influence of these tools on the Honegumi workflow are described below and presented visually in \cref{fig:flowchart}.

When a researcher visits the website, an HTML script runs that downloads PyScript and the most recent version of the \texttt{honegumi} package that has passed a unit testing framework. The use of unit testing ensures that updates to the code base do not break one of the 4096 possible scripts generated by selection combinations. Once loaded, PyScript calls the \texttt{honegumi} API on a default selection configuration. The \texttt{honegumi} API uses Jinja to render a Jinja template file into a Bayesian optimization script, which is then passed back to PyScript to be displayed to the user directly beneath the selection grid (\cref{fig:selection_interface}). The Jinja template file encodes the logic around which lines should be added, removed, or modified based on the selected set of options. Likewise, the code that renders the Jinja template encodes incompatible feature combinations, either due to not being implemented or being logically inconsistent. These incompatible configurations are crossed out. The act of pulling appropriate coding templates and displaying them to the user is repeated each time a new configuration option is selected.

The following subsections provide additional technical details on the implementations of the key tools enabling the framework.

\subsection{Jinja Templates}

The BO scripts presented to the user are generated using the Jinja templating engine. Jinja allows a suite of conditional "if" statements to map user selected options to a resulting script. These are sufficiently expressive to allow for multi-conditioned "if" statements to be used in cases where several options might create conflicts within the generated script. 

For example, take the case where one wants to switch between a function that returns a single output (single-objective) and one with two outputs (multi-objective) depending on the value of \texttt{obj}. The code blocks below show the Jinja template and the results depending on the value of \texttt{obj}:\\

\begin{lstlisting}[
  language=Python, 
  label={lst:jinja}
]
def fn{% if obj == "multi" %}_moo{% endif %}(x1, x2):

    y = 0.5*x1 + 0.2*x2
    
    {% if obj == "multi" -%}
    y2 = 0.2*x1 + 0.5*x2
    
    return {obj1_name: y, obj2_name: y2}
    {% else %} {# single objective #}

    return y
    {%- endif %}
\end{lstlisting}

% If this Jinja template were to be rendered with \texttt{obj="single"}, then the following single-objective script would be produced:

\noindent If \texttt{obj = "single"}, Jinja renders:

\begin{lstlisting}[
  language=Python, 
  label={lst:jinja-single}
]
def fn(x1, x2):
    y = 0.5*x1 + 0.2*x2
    return y

\end{lstlisting}

\noindent If \texttt{obj = "multi"}, Jinja renders:

\begin{lstlisting}[
  language=Python,  
  label={lst:jinja-multi}
]
def fn_moo(x1, x2):
    y = 0.5*x1 + 0.2*x2
    y2 = 0.2*x1 + 0.5*x2
    return {obj1_name: y, obj2_name: y2}
\end{lstlisting}

The full implementation is available at \href{https://github.com/sgbaird/honegumi/blob/main/src/honegumi/ax/main.py.jinja}{main.py.jinja} [\href{https://github.com/sgbaird/honegumi/blob/5ea978626c8dcd5869df1e4032ef837574c1e3fb/src/honegumi/ax/main.py.jinja}{permalink}] within the GitHub repository. Jinja is also used to create the HTML file with the Honegumi selection grid via \href{https://github.com/sgbaird/honegumi/blob/main/src/honegumi/core/honegumi.html.jinja}{honegumi.html.jinja} [\href{https://github.com/sgbaird/honegumi/blob/5ea978626c8dcd5869df1e4032ef837574c1e3fb/src/honegumi/core/honegumi.html.jinja}{permalink}].

% https://github.com/sgbaird/honegumi/blob/b2ae0275e02452162b00654fd92b03348bb46af7/src/honegumi/ax/main.py.jinja

% https://github.com/sgbaird/honegumi/blob/f963411997acaed046ad9e3f17f14f8ee3bbefd3/src/honegumi/core/honegumi.html.jinja

\subsection{PyScript}

Honegumi functions principally off of PyScript, which allows Python to be run in the browser and Honegumi to have a similar feel to a web application. Upon initialization, PyScript and Honegumi are installed. PyScript executes a top-level function within the \texttt{honegumi} package using the current grid selections as inputs, and outputs the rendered BO script. Each time a new selection is made, PyScript runs the function again.

Alternatively, we could have written all of the Honegumi functionality in pure Javascript, or taken a separate approach of hosting Honegumi as a dedicated web app with a full Python environment. Rather than splitting between languages, a Python-based approach was preferred for maintainability purposes. However, creating a dedicated web app with a Python environment would reduce the portability and separate the interface from the documentation site. PyScript offered a straightforward middle path where most of the development could still be done in Python and the interface could still be directly embedded within a static website.

Since PyScript uses WebAssembly and MicroPython, it can be run on most browsers; however, one limitation is that the Python code and dependencies must be based on pure Python. This is not a problem, since Honegumi and all dependencies are written this way.

\section{Conclusion}

Honegumi represents a new approach to increasing the accessibility of Bayesian optimization tools in experimental research. By combining an interactive interface with educational resources, it addresses both the technical and conceptual barriers that researchers often face when implementing BO in their work. The selection grid allows researchers to quickly generate working BO scripts while the accompanying tutorials provide the knowledge needed to modify these scripts appropriately for specific experimental contexts. This integration of practical tools with educational content helps researchers not only implement BO methods but also understand the principles behind their choices.

Looking forward, Honegumi's modular structure creates opportunities for expansion to other optimization platforms and experimental domains. The framework demonstrated here - using interactive interfaces to simplify complex computational tools while simultaneously teaching their proper use - may serve as a model for making other advanced computational methods more accessible to experimental researchers. As computational techniques become increasingly central to scientific discovery, tools that bridge the gap between theoretical capabilities and practical implementation will play an important role in accelerating research progress.

% \begin{itemize}
%     \item tool for the research community to accelerate the inclusion of more advanced tools for guiding experimental design
%     \item tool not only provides scripts, but teaches people how to use it such that they can interact with the underlying code base more easily going forward.
%     \item Modularity may allow other tools to adopt
%     \item example of how interfaces change how researchers can interact with complicated code bases and how such tools can be pedagogical with minimal effort.
% \end{itemize}

% \bibliographystyle{elsarticle-num-names}
\bibliographystyle{rsc}
\bibliography{references}

\providecommand*{\mcitethebibliography}{\thebibliography}
\csname @ifundefined\endcsname{endmcitethebibliography}
{\let\endmcitethebibliography\endthebibliography}{}
\begin{mcitethebibliography}{12}
\providecommand*{\natexlab}[1]{#1}
\providecommand*{\mciteSetBstSublistMode}[1]{}
\providecommand*{\mciteSetBstMaxWidthForm}[2]{}
\providecommand*{\mciteBstWouldAddEndPuncttrue}
  {\def\EndOfBibitem{\unskip.}}
\providecommand*{\mciteBstWouldAddEndPunctfalse}
  {\let\EndOfBibitem\relax}
\providecommand*{\mciteSetBstMidEndSepPunct}[3]{}
\providecommand*{\mciteSetBstSublistLabelBeginEnd}[3]{}
\providecommand*{\EndOfBibitem}{}
\mciteSetBstSublistMode{f}
\mciteSetBstMaxWidthForm{subitem}
{(\emph{\alph{mcitesubitemcount}})}
\mciteSetBstSublistLabelBeginEnd{\mcitemaxwidthsubitemform\space}
{\relax}{\relax}

\bibitem[Strieth-Kalthoff \emph{et~al.}(2024)Strieth-Kalthoff, Hao, Rathore, Derasp, Gaudin, Angello, Seifrid, Trushina, Guy, Liu, Tang, Mamada, Wang, Tsagaantsooj, Lavigne, Pollice, Wu, Hotta, Bodo, Li, Haddadnia, Wołos, Roszak, Ser, Bozal-Ginesta, Hickman, Vestfrid, Aguilar-Granda, Klimareva, Sigerson, Hou, Gahler, Lach, Warzybok, Borodin, Rohrbach, Sanchez-Lengeling, Adachi, Grzybowski, Cronin, Hein, Burke, and Aspuru-Guzik]{strieth-kalthoff_delocalized_2024}
F.~Strieth-Kalthoff, H.~Hao, V.~Rathore, J.~Derasp, T.~Gaudin, N.~H. Angello, M.~Seifrid, E.~Trushina, M.~Guy, J.~Liu, X.~Tang, M.~Mamada, W.~Wang, T.~Tsagaantsooj, C.~Lavigne, R.~Pollice, T.~C. Wu, K.~Hotta, L.~Bodo, S.~Li, M.~Haddadnia, A.~Wołos, R.~Roszak, C.~T. Ser, C.~Bozal-Ginesta, R.~J. Hickman, J.~Vestfrid, A.~Aguilar-Granda, E.~L. Klimareva, R.~C. Sigerson, W.~Hou, D.~Gahler, S.~Lach, A.~Warzybok, O.~Borodin, S.~Rohrbach, B.~Sanchez-Lengeling, C.~Adachi, B.~A. Grzybowski, L.~Cronin, J.~E. Hein, M.~D. Burke and A.~Aspuru-Guzik, \emph{Science}, 2024, \textbf{384}, eadk9227\relax
\mciteBstWouldAddEndPuncttrue
\mciteSetBstMidEndSepPunct{\mcitedefaultmidpunct}
{\mcitedefaultendpunct}{\mcitedefaultseppunct}\relax
\EndOfBibitem
\bibitem[Shields \emph{et~al.}(2021)Shields, Stevens, Li, Parasram, Damani, Alvarado, Janey, Adams, and Doyle]{shields2021bayesian}
B.~J. Shields, J.~Stevens, J.~Li, M.~Parasram, F.~Damani, J.~I.~M. Alvarado, J.~M. Janey, R.~P. Adams and A.~G. Doyle, \emph{Nature}, 2021, \textbf{590}, 89--96\relax
\mciteBstWouldAddEndPuncttrue
\mciteSetBstMidEndSepPunct{\mcitedefaultmidpunct}
{\mcitedefaultendpunct}{\mcitedefaultseppunct}\relax
\EndOfBibitem
\bibitem[Agarwal \emph{et~al.}(2021)Agarwal, Doan, Robertson, Zhang, and Assary]{agarwal2021discovery}
G.~Agarwal, H.~A. Doan, L.~A. Robertson, L.~Zhang and R.~S. Assary, \emph{Chemistry of Materials}, 2021, \textbf{33}, 8133--8144\relax
\mciteBstWouldAddEndPuncttrue
\mciteSetBstMidEndSepPunct{\mcitedefaultmidpunct}
{\mcitedefaultendpunct}{\mcitedefaultseppunct}\relax
\EndOfBibitem
\bibitem[Chen \emph{et~al.}(2020)Chen, Wang, Li, Hou, and Yin]{chen2020bayesian}
X.~Chen, C.~Wang, Z.~Li, Z.~Hou and W.-J. Yin, \emph{Science China Materials}, 2020,  1024--1035\relax
\mciteBstWouldAddEndPuncttrue
\mciteSetBstMidEndSepPunct{\mcitedefaultmidpunct}
{\mcitedefaultendpunct}{\mcitedefaultseppunct}\relax
\EndOfBibitem
\bibitem[Hickman \emph{et~al.}(2022)Hickman, Aldeghi, H{\"a}se, and Aspuru-Guzik]{hickman2022bayesian}
R.~J. Hickman, M.~Aldeghi, F.~H{\"a}se and A.~Aspuru-Guzik, \emph{Digital Discovery}, 2022, \textbf{1}, 732--744\relax
\mciteBstWouldAddEndPuncttrue
\mciteSetBstMidEndSepPunct{\mcitedefaultmidpunct}
{\mcitedefaultendpunct}{\mcitedefaultseppunct}\relax
\EndOfBibitem
\bibitem[Arr{\'o}yave \emph{et~al.}(2022)Arr{\'o}yave, Khatamsaz, Vela, Couperthwaite, Molkeri, Singh, Johnson, Qian, Srivastava, and Allaire]{arroyave2022perspective}
R.~Arr{\'o}yave, D.~Khatamsaz, B.~Vela, R.~Couperthwaite, A.~Molkeri, P.~Singh, D.~D. Johnson, X.~Qian, A.~Srivastava and D.~Allaire, \emph{MRS communications}, 2022, \textbf{12}, 1037--1049\relax
\mciteBstWouldAddEndPuncttrue
\mciteSetBstMidEndSepPunct{\mcitedefaultmidpunct}
{\mcitedefaultendpunct}{\mcitedefaultseppunct}\relax
\EndOfBibitem
\bibitem[Tom \emph{et~al.}(2024)Tom, Schmid, Baird, Cao, Darvish, Hao, Lo, Pablo-Garc{\'\i}a, Rajaonson, Skreta,\emph{et~al.}]{tom2024self}
G.~Tom, S.~P. Schmid, S.~G. Baird, Y.~Cao, K.~Darvish, H.~Hao, S.~Lo, S.~Pablo-Garc{\'\i}a, E.~M. Rajaonson, M.~Skreta \emph{et~al.}, \emph{Chemical Reviews}, 2024, \textbf{124}, 9633--9732\relax
\mciteBstWouldAddEndPuncttrue
\mciteSetBstMidEndSepPunct{\mcitedefaultmidpunct}
{\mcitedefaultendpunct}{\mcitedefaultseppunct}\relax
\EndOfBibitem
\bibitem[Frazier(2018)]{frazier_tutorial_2018}
P.~I. Frazier, \emph{A {Tutorial} on {Bayesian} {Optimization}}, 2018, \url{https://arxiv.org/abs/1807.02811v1}\relax
\mciteBstWouldAddEndPuncttrue
\mciteSetBstMidEndSepPunct{\mcitedefaultmidpunct}
{\mcitedefaultendpunct}{\mcitedefaultseppunct}\relax
\EndOfBibitem
\bibitem[Garnett(2023)]{garnett_bayesian_2023}
R.~Garnett, \emph{Bayesian {Optimization}}, Cambridge University Press, 2023\relax
\mciteBstWouldAddEndPuncttrue
\mciteSetBstMidEndSepPunct{\mcitedefaultmidpunct}
{\mcitedefaultendpunct}{\mcitedefaultseppunct}\relax
\EndOfBibitem
\bibitem[Shahriari \emph{et~al.}(2016)Shahriari, Swersky, Wang, Adams, and de~Freitas]{shahriari_taking_2016}
B.~Shahriari, K.~Swersky, Z.~Wang, R.~P. Adams and N.~de~Freitas, \emph{Proceedings of the IEEE}, 2016, \textbf{104}, 148--175\relax
\mciteBstWouldAddEndPuncttrue
\mciteSetBstMidEndSepPunct{\mcitedefaultmidpunct}
{\mcitedefaultendpunct}{\mcitedefaultseppunct}\relax
\EndOfBibitem
\bibitem[Balandat \emph{et~al.}(2020)Balandat, Karrer, Jiang, Daulton, Letham, Wilson, and Bakshy]{balandat_botorch_2020}
M.~Balandat, B.~Karrer, D.~Jiang, S.~Daulton, B.~Letham, A.~G. Wilson and E.~Bakshy, Advances in {Neural} {Information} {Processing} {Systems}, 2020, pp. 21524--21538\relax
\mciteBstWouldAddEndPuncttrue
\mciteSetBstMidEndSepPunct{\mcitedefaultmidpunct}
{\mcitedefaultendpunct}{\mcitedefaultseppunct}\relax
\EndOfBibitem
\bibitem[Gardner \emph{et~al.}(2018)Gardner, Pleiss, Weinberger, Bindel, and Wilson]{gardner_gpytorch_2018}
J.~Gardner, G.~Pleiss, K.~Q. Weinberger, D.~Bindel and A.~G. Wilson, Advances in {Neural} {Information} {Processing} {Systems}, 2018\relax
\mciteBstWouldAddEndPuncttrue
\mciteSetBstMidEndSepPunct{\mcitedefaultmidpunct}
{\mcitedefaultendpunct}{\mcitedefaultseppunct}\relax
\EndOfBibitem
\end{mcitethebibliography}

\end{document}